\documentclass{article}
\usepackage{tipa}
\usepackage{pwpl_style}
\usepackage{natbib}
\usepackage{multirow}
\usepackage{graphicx}

\begin{document}

\title{Investigating the Sensitivity of Automatic Speech Recognition \\Systems to Phonetic Variation in L2 Englishes}

\author{Emma O'Neill \and Julie Carson-Berndsen\thanks{This research was conducted with the financial support of the School of Computer Science at the ADAPT SFI Research Centre at University College Dublin. ADAPT, the SFI Research Centre for AI-Driven Digital Content Technology, is funded by Science Foundation Ireland through the SFI Research Centres Programme.}} 

\maketitle
\shorttitle{Investigating the sensitivity of ASR Systems to Phonetic Variation}

\section{Introduction}

It is generally agreed that, whilst Automatic Speech Recognition (ASR) systems are capable of delivering ceiling-level performance on so-called ``standard" speech, this performance typically deteriorates when it comes to spoken varieties that are underrepresented or absent in the data used to train such systems. With the complex network architectures and ``black box" nature of many high performing ASR systems, it is often difficult to pinpoint exactly which features of a spoken language variety prompt such performance deterioration beyond just deviation from the expected ``norm". In this work, a method of analysing how one such ASR system handles pronunciation variation at the phonetic level is explored and more specifically, how particular phonetic realisations lead to the misrecognition of phonemes. 
It is demonstrated that the erroneous ASR output is systematic and consistent across speakers with similar spoken varieties. Furthermore, it is examined how these misrecognitions by the ASR compare with the observations of human annotators. This has real world significance both for the development and improvement of ASR systems that are capable of handling different language varieties and for the use of ASR technology as a tool in annotation and variation analysis.
\section{Background}

Most work involving spoken data, whether it be for the purposes of developing language technology or (socio)linguistic research and analysis, requires high quality transcripts of the speech. Typically, audio data is transcribed at the orthographic or word level but, depending upon the use case, datasets may also be annotated for features such as sentiment, syntax, or phonetic realisations. Manual transcription, however, is a time and resource heavy process. As such, it is becoming increasingly popular to consider using ASR as a potential step in the annotation pipeline, typically by generating automated transcripts which can then be manually corrected by human annotators. For example, DARLA \citep{reddy2015toward} is a linguistic tool which makes use of ASR, with its BedWord service providing fully automated transcripts of audio data. \citet{jimerson2018asr} investigate the development of ASR systems for under-resourced languages to aid in their documentation and preservation and \citet{michaud2018integrating} discuss the success in integrating ASR into the language documentation workflow and the benefits for linguists. \citet{omachi2021end} propose an end-to-end ASR model which simultaneously transcribes speech orthographically and provides linguistic annotations such as phoneme transcripts and part-of-speech tags. 

Despite the obvious benefits to automating, at least partially, the annotation process, the incorporation of ASR has yet to become widespread in the field. This is, in part, because it has been well established that ASR systems exhibit comparatively poorer performance with some speakers than others. \citet{koenecke2020racial} investigated the commercial ASR systems of Amazon, Apple, Google, IBM, and Microsoft and found significantly higher word error rates (WER) for African American varieties of English. This deterioration of performance was attributed to underrepresentation in the training data of the system and the authors call for data collection to include ``nonstandard varieties [...] including those with regional and nonnative-English accents". Race and dialect were also shown to impact the accuracy of the automatic Youtube captions in work by \citet{tatman2017effects} where it was noted that the best performance was achieved on white speakers with a General American (non-regional) dialect. 
In an investigation into the use of ASR for the purposes of linguistic annotation, \citet{markl2022commercial} notes the potential time-saving benefits to incorporating ASR particularly for large-scale data sets. However, they outline the limitations of such tools and the challenges faced in applying ASR to diverse speech datasets containing various accents, topics, and recording environments. This work also points out that the manual correction of automated transcripts can be challenging and encourages collaboration with speech technology experts to determine the most appropriate tool for an annotation project. 

It seems clear, then, that the incorporation of ASR technology could be of great benefit to large-scale linguistic research. However, the deterioration of performance when used with ``non-standard" speech is a barrier for many researchers. Even when choosing an ASR solution best suited for the speech data to be annotated, the manual correction of any errors in the output can still be a difficult task when the behaviour of the system is not fully understood. As such, this work analyses the performance of an ASR system at the phonetic level for two main reasons. Firstly, by discovering which pronunciation variants cause problems during recognition, a focused and targeted approach can be applied to the collection of additional data for fine-tuning thus improving the ASR performance for minimal cost. Secondly, a fuller understanding of how an ASR systems handles a particular spoken variety and the errors it is likely to make can potentially ease the difficulty of manually correcting the automated transcripts.
\section{Methods}
\subsection{Data}

The audio data used in this experiment is taken from the L2 Arctic Corpus \citep{zhao2018l2}. This corpus contains recordings of 24 non-native speakers of English across six L1s, namely Arabic, Hindi, Korean, Mandarin, Spanish, and Vietnamese. For each L1 there are four speakers, two male and two female. Whilst each speaker recorded approximately one hour of read speech using the Arctic prompts \citep{kominek2004cmu}, this work focuses on the subset of recordings which are manually annotated at the phonemic level for insertions, deletions, and substitutions. A breakdown of the number of utterances recorded for each speaker and the total time can be seen in Table \ref{tab:l2_speakers}.

\begin{table}[t]
\centering
\resizebox{0.8\textwidth}{!}{%
\begin{tabular}{|lcc|c|c|}
\hline
\multicolumn{1}{|l|}{\textbf{Speaker}} & \multicolumn{1}{l|}{\textbf{L1}} & \textbf{Gender} & \textbf{No. of utterances} & \textbf{Minutes of speech} \\ \hline
\multicolumn{1}{|l|}{ABA}     & \multicolumn{1}{l|}{Arabic}     & M      & 150               & 10.32             \\ 
\multicolumn{1}{|l|}{SKA}     & \multicolumn{1}{l|}{Arabic}     & F      & 150               & 8.21              \\ 
\multicolumn{1}{|l|}{YBAA}    & \multicolumn{1}{l|}{Arabic}     & M      & 149               & 9.10              \\ 
\multicolumn{1}{|l|}{ZHAA}    & \multicolumn{1}{l|}{Arabic}     & F      & 150               & 8.38              \\ 
\multicolumn{1}{|l|}{BWC}     & \multicolumn{1}{l|}{Mandarin}   & M      & 150               & 10.62             \\ 
\multicolumn{1}{|l|}{LXC}     & \multicolumn{1}{l|}{Mandarin}   & F      & 150               & 9.64              \\ 
\multicolumn{1}{|l|}{NCC}     & \multicolumn{1}{l|}{Mandarin}   & F      & 150               & 9.35              \\ 
\multicolumn{1}{|l|}{TXHC}    & \multicolumn{1}{l|}{Mandarin}   & M      & 150               & 8.73              \\ 
\multicolumn{1}{|l|}{ASI}     & \multicolumn{1}{l|}{Hindi}      & M      & 150               & 7.51              \\ 
\multicolumn{1}{|l|}{RRBI}    & \multicolumn{1}{l|}{Hindi}      & M      & 150               & 8.69              \\ 
\multicolumn{1}{|l|}{SVBI}    & \multicolumn{1}{l|}{Hindi}      & F      & 150               & 7.06              \\ 
\multicolumn{1}{|l|}{TNI}     & \multicolumn{1}{l|}{Hindi}      & F      & 150               & 8.32              \\ 
\multicolumn{1}{|l|}{HJK}     & \multicolumn{1}{l|}{Korean}     & F      & 150               & 7.54              \\ 
\multicolumn{1}{|l|}{HKK}     & \multicolumn{1}{l|}{Korean}     & M      & 150               & 8.91              \\ 
\multicolumn{1}{|l|}{YDCK}    & \multicolumn{1}{l|}{Korean}     & F      & 150               & 10.17             \\ 
\multicolumn{1}{|l|}{YKWK}    & \multicolumn{1}{l|}{Korean}     & M      & 150               & 8.81              \\ 
\multicolumn{1}{|l|}{EBVS}    & \multicolumn{1}{l|}{Spanish}    & M      & 150               & 9.91              \\ 
\multicolumn{1}{|l|}{ERMS}    & \multicolumn{1}{l|}{Spanish}    & M      & 150               & 10.44             \\ 
\multicolumn{1}{|l|}{MBMPS}   & \multicolumn{1}{l|}{Spanish}    & F      & 150               & 11.52             \\ 
\multicolumn{1}{|l|}{NJS}     & \multicolumn{1}{l|}{Spanish}    & F      & 150               & 8.09              \\ 
\multicolumn{1}{|l|}{HQTV}    & \multicolumn{1}{l|}{Vietnamese} & M      & 150               & 9.20              \\ 
\multicolumn{1}{|l|}{PNV}     & \multicolumn{1}{l|}{Vietnamese} & F      & 150               & 9.53              \\ 
\multicolumn{1}{|l|}{THV}     & \multicolumn{1}{l|}{Vietnamese} & F      & 150               & 9.44              \\ 
\multicolumn{1}{|l|}{TLV}     & \multicolumn{1}{l|}{Vietnamese} & M      & 150               & 10.32             \\ \hline
\multicolumn{3}{|c|}{Total}                                              & 3599              & 219.81            \\ \hline
\end{tabular}}
\caption{Speaker breakdown of the annotated subset of the L2-Arctic Corpus.}
\label{tab:l2_speakers}
\end{table}

\subsection{ASR}

A pretrained wav2vec 2.0 model \citep{baevski2020wav2vec} was used as the ASR component of this work. 
This is a transformer based model that first learns latent speech representations in a self-supervised fashion using unlabelled data and is then fine-tuned on 100 hours of labelled data from the Librispeech corpus \citep{panayotov2015librispeech}. 
Given the nature of the training data, this model is most suited to North American varieties of English. Since this model does not use a word-level language model component when generating output, it is able to produce text representations of audio which contain non-words and unlikely word sequences. These outputs often better capture the phonetic sequence of the speech. An example of this can be seen in Example (1) where the prompt text (a) was read by an L1 Mandarin speaker and produced the ASR transcript (b) which would appear to capture common phonetic features of Mandarin Accented English including vowel epenthesis and difficulties with the tense/lax distinction \citep{siqi2012phonological, broselow1998emergence, nogita2012not, huang2014revisiting, khanal2021mispronunciation}.
\\
\\
\indent (1)  a. It was \textbf{simple}\hspace{1pt} in its way and \hspace{5pt} no virtue of \textbf{his} \\
\indent\hspace{12pt}  b. It was \textbf{simbol} in its way and \textbf{a} no virtue of \textbf{ease}

\subsection{Phoneme Sequence Alignment}

The prompt texts and ASR transcripts were converted to their corresponding phoneme sequences in ARPAbet notation using the CMU Pronouncing Dictionary \citep{weide1998cmu} and, for non-words, a grapheme-to-phoneme tool trained on this dictionary \citep{g2p2016tool}. 
This method was chosen for two reasons. Firstly, the annotated portion of the L2-Arctic corpus also uses phoneme sequences from the CMU Pronunciation Dictionary and so direct comparisons are possible in the forthcoming analysis. Furthermore, since the pronunciations in this dictionary are typically suited to North American varieties of English, they capture the likely pronunciation of the speech which the ASR model was trained on. Thus, whilst these phoneme sequences for the words in the prompt text are treated as a `ground truth' when investigating phoneme insertions, deletions, and substitutions, it is not suggested that these are the `correct' pronunciations but, rather, the pronunciations which are `expected' by the ASR.

In order to align the phoneme sequences of the ASR transcript and of the prompt text, a weighted-edit distance measure and alignment algorithm akin to that of \citet{wagner1974string} was employed. Typically, when finding the optimal alignment between two strings, the Levenshtein edit-distance \citep{levenshtein1966binary} is used. This is a measure of the number of edit operations (insertions, deletions, and substitutions) required to transform one sequence into another. However, this measure only considers the total number of edit operations and treats each as being equally likely and having equal cost which can lead to unintuitive alignments. Consider, for instance, the final words in the ASR transcript and prompt text of the previous example. The word ``ease" is represented phonemically as /IY Z/, whilst ``his" is represented as /HH IH Z/. Using the Levenshtein edit-distance measure, both the alignments depicted in Table \ref{tab:levenshtein} would be equally likely. The first alignment is where the /HH/ in the target word was substituted with /IY/ and the subsequent /IH/ was deleted whilst the second alignment suggests the /HH/ was deleted and the /IH/ was substituted with /IY/. 

\begin{table}[t]
\centering
\begin{tabular}{|l|lll|lll|}
\hline
                & \multicolumn{3}{l|}{\textbf{Alignment 1}} & \multicolumn{3}{l|}{\textbf{Alignment 2}} \\ \hline
``his''         & HH         & IH        & Z       & HH         & IH        & Z       \\
``ease''        & IY         &           & Z       &            & IY        & Z       \\
Edit Operations & sub        & del       &         & del        & sub       &         \\ \hline
\end{tabular}
\caption{Two equally likely alignment options using the Levenshtein edit-distance measure.}
\label{tab:levenshtein}
\end{table}

\begin{figure}[h]
    \centering
    \includegraphics[width=\linewidth]{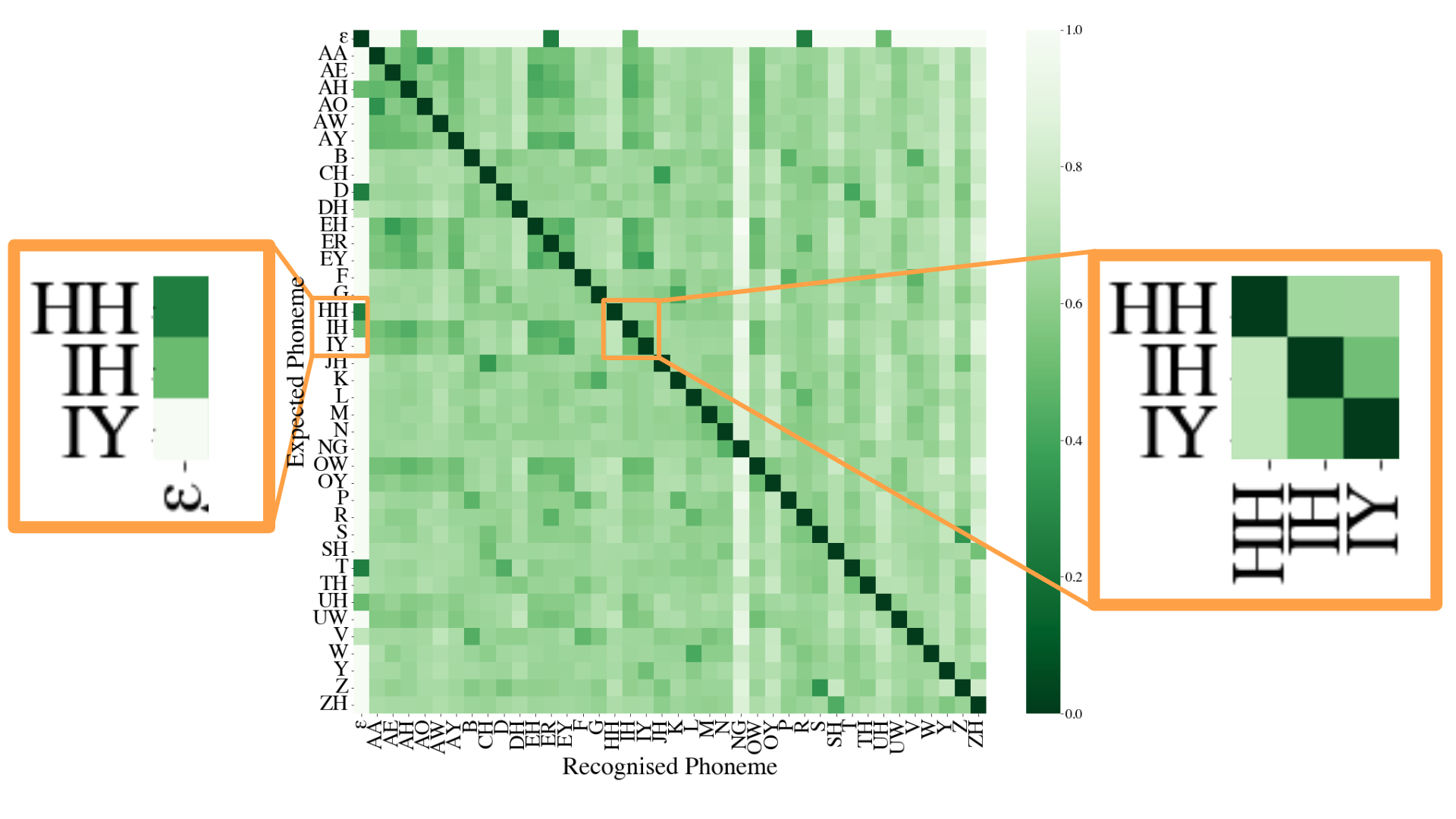}
    \caption{Weight matrix used in the phoneme alignment algorithm visualised as a heatmap.}
    \label{fig:heatmap}
\end{figure}

Given the much higher likelihood of the elision of the glottal fricative /HH/ and the tensing of the high front vowel /IH/ to /IY/, Alignment 2 is preferred from a variation analysis standpoint. To reliably achieve these more probable alignments, a weighted edit-distance measure was instead used whereby more likely operations carried a lower cost during alignment and were thus preferred. The weights associated with each possible edit operation were taken from a phoneme similarity matrix developed previously based on the acoustic and distributional properties of the phonemes and their involvement in epenthesis and elision processes. For full details, see \citealp{o2019effect} and \citealt{o2020s}. The resultant weight matrix, visualised as a heatmap for clarity, can be seen in Figure \ref{fig:heatmap}. Here, darker cells reflect more similar phonemes and thus lower substitution costs. Insertions and deletions are treated as substitutions involving the empty string ($\varepsilon$). Thus, it can be seen that /HH/ is more likely to be deleted than /IH/ and the /IH/-/IY/ substitution is more probable than /HH/-/IY/.

After alignment, counts of all possible insertions, deletions, and substitutions are calculated for each speaker. This allows us to determine whether speakers of the same L1, who presumably produce similar patterns of variation in their spoken English, result in similar error patterns in their ASR transcripts. We are also able to directly compare the recognition rates and common substitutes as determined by the ASR transcripts with judgements from human annotators. 

\section{Results}
\subsection{Speaker Level Confusion Matrices}

\begin{figure}[ht]
    \centering
    \includegraphics[width=0.95\linewidth]{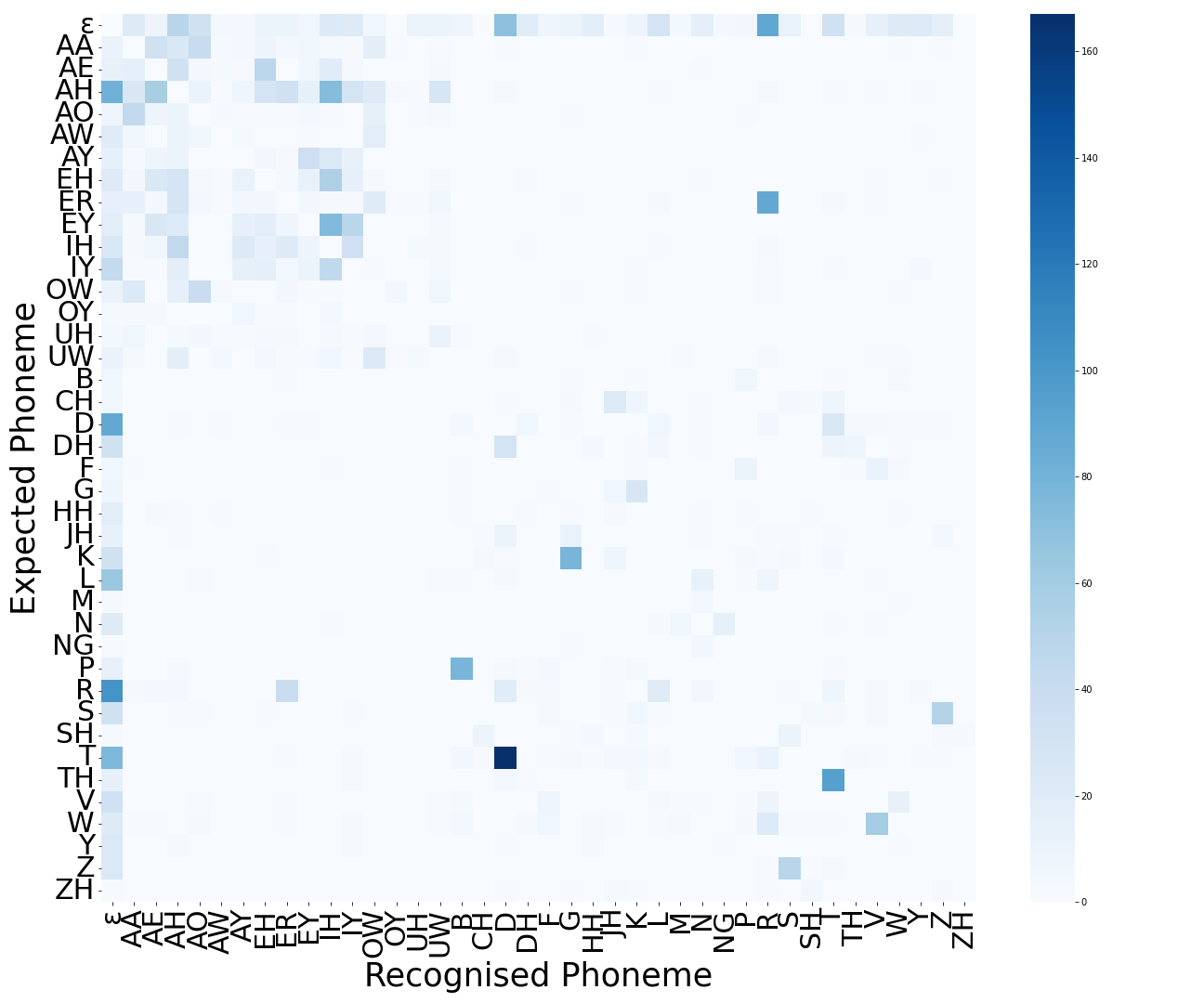}
    \caption{Confusion matrix generated from the ASR transcripts from an L1 Hindi speaker.}
    \label{fig:hindi}
\end{figure}

After alignment, counts of the observed edit operations were stored in a confusion matrix for each speaker where, as with the previously discussed similarity matrix, insertions and deletions are treated as substitutions with the empty string ($\varepsilon$). 
Figure \ref{fig:hindi} depicts one such confusion matrix (again visualised as a heatmap for clarity) for an L1 Hindi speaker. In this case, darker cells indicate higher counts of a particular substitution. Here, the highest value and most common substitution across all of the read prompts from this speaker is in the case of an expected /T/ phoneme being represented in the ASR transcript as a /D/ phoneme. This observed substitution stems from the speaker's realisation of /T/ phonemes as either the retroflex [\textipa{\textrtailt}] or unaspirated [t] in word initial position. Since the ASR model is unfamiliar with such productions, it typically mistakes them for instances of the /D/ phoneme instead thus resulting in high substitution counts and a deterioration of the word error rate. 

\subsection{Speaker Clustering in Multi-Dimensional Space}

The speaker level confusion matrices reveal features of variation in the pronunciation of a single speaker, highlighting weaknesses of the ASR and gaps in its training data. It can be surmised, from the previous example, that more instances of retroflex [\textipa{\textrtailt}] in the training and fine-tuning data would improve the performance of the ASR for that speaker. More significantly, if many speakers exhibit similar patterns of pronunciation variation, and the ASR performs similarly across these speakers, then this addition to the training data would improve the recognition performance for a much larger group of speakers. To investigate this idea, it needs to be determined whether or not the ASR performance is consistent and systematic across speakers with similar varieties. 

\begin{figure}[t]
    \centering
    \includegraphics[width=0.85\linewidth]{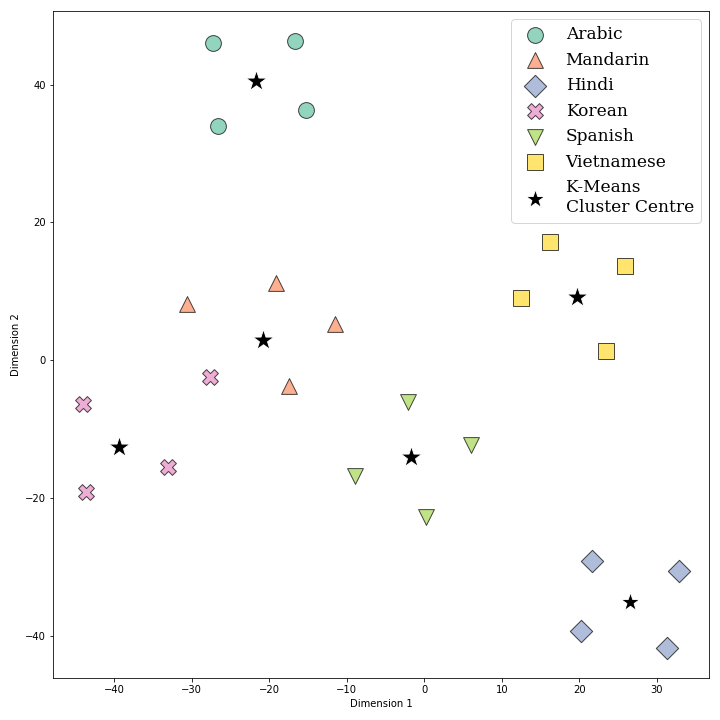}
    \caption{2-dimensional visualisation of speaker representations and cluster centers.}
    \label{fig:speaker_vis}
\end{figure}

If it were the case that the ASR produces systematic misrecognitions at the phoneme level for speakers who exhibit similar patterns of variation, it would be expected that speakers with the same L1, who likely share features of pronunciation, result in similar confusion matrices using the method described previously. The ASR performance for each speaker is essentially represented by the 40x40 confusion matrix, the values of which can be reshaped into a vector of 1600 dimensions. In theory, if two speakers produced similar confusion matrices, then there would be a relatively small distance between their vector representations when plotted in multi-dimensional space. With this in mind, 
the confusion matrices for each of the 24 speakers were reshaped into 1600 dimensional vectors and a k-means clustering approach \citep{macqueen1967classification} was used to determine whether speakers of the same L1 were typically located in the same area in the multi-dimensional space. K-means clustering is an unsupervised machine learning approach which endeavours to find the centers of a predefined number of clusters within the data points. In this case, 6 clusters are used to match the number of distinct L1s in the corpus. Whilst the clustering was carried out in 1600 dimensions, t-SNE \citep{van2008visualizing} was used to visualise the speaker representations and cluster centers in 2 dimensions. This can be seen in Figure \ref{fig:speaker_vis} which demonstrates how speakers of the same L1 do tend to cluster together in the space and so have similar vector representations resulting from similar confusion matrices.

\subsection{Comparison with Human Annotator Judgements}

Having established that the ASR output produces patterns of errors in relation to systematic pronunciation variation across speakers with similar varieties, a more fine-grained analysis of its performance with specific phonetic variation is carried out through comparison with human annotator judgements. As discussed previously, the audio data used to construct the speaker level confusion matrices was the subset of the L2 Arctic corpus that was manually annotated at the phoneme level for substitutions, insertions, and deletions. Thus, we are able to compare the frequency with which a phoneme is realised in such a way that the ASR produces erroneous output with the frequency in which a human annotator believes the phoneme to have been substituted for another. Furthermore, the specific phoneme substitutions, as determined by the ASR output, can be compared with the phoneme substitutions identified by the annotator to determine whether the ASR output captures the same judgements as made by a human listener. A summary of this comparison is given in Table \ref{tab:ha_comparison}.

\begin{table}[ht]
\resizebox{\textwidth}{!}{%
\begin{tabular}{|c|c|c|c|c|c|c|c|}
\hline
\multicolumn{1}{|c|}{\textbf{L1}} & \multicolumn{1}{c|}{\textbf{\begin{tabular}[c]{@{}c@{}}Target\\ Phoneme\end{tabular}}} & \multicolumn{1}{c|}{\textbf{\begin{tabular}[c]{@{}c@{}}Recognition\\ Rate\\ (ASR)\end{tabular}}} & \multicolumn{1}{c|}{\textbf{\begin{tabular}[c]{@{}c@{}}Recognition\\ Rate\\ (HA)\end{tabular}}} & \multicolumn{1}{c|}{\textbf{\begin{tabular}[c]{@{}c@{}}Most\\ Common\\ Substitute\\ (ASR)\end{tabular}}} & \multicolumn{1}{c|}{\textbf{\begin{tabular}[c]{@{}c@{}}Most\\ Common\\ Substitute\\ (HA)\end{tabular}}} & \multicolumn{1}{c|}{\textbf{\begin{tabular}[c]{@{}c@{}}MCS \\ Substitution\\ Rate\\ (ASR)\end{tabular}}} & \multicolumn{1}{c|}{\textbf{\begin{tabular}[c]{@{}c@{}}MCS \\ Substitution\\ Rate\\ (HA)\end{tabular}}} \\ \hline

\multirow{3}{*}{Arabic}
& /EH/ & 74.8\% & 89.0\% & /IH/ & /IH/ & 13.9\% & 4.1\% \\
& /P/  & 76.0\% & 71.4\% & /B/  & /B/  & 21.6\% & 25.8\%  \\
& /TH/ & 79.2\% & 81.0\% & /S/  & /S/  & 7.5\%  & 13.1\% \\ \hline
\multirow{3}{*}{Mandarin}
& /TH/ & 70.2\% & 50.9\% & /S/  & /S/  & 12.9\% & 37.9\%   \\
& /ZH/ & 70.8\% & 62.5\% & /SH/ & /SH/ & 8.3\% & 29.2\%  \\
& /UH/ & 71.7\% & 89.3\% & /UW/ & /UW/ & 5.0\% & 6.6\% \\ \hline
\multirow{3}{*}{Hindi}
& /OY/ & 57.5\% & 95.0\% & /AY/ & /AY/ & 25.0\% & 5.0\% \\
& /AW/ & 69.7\% & 79.3\% & /OW/ & /AO/ & 15.9\% & 8.6\% \\
& /TH/ & 71.8\% & 41.4\% & /T/  & /T/  & 21.1\% & 45.9\%  \\ \hline
\multirow{3}{*}{Korean}     
& /ZH/ & 60.0\% & 47.4\% & /Z/  & /JH/ & 15.0\% & 26.3\% \\
& /TH/ & 75.0\% & 75.4\% & /S/  & /S/  & 12.1\% & 15.3\% \\
& /JH/ & 78.6\% & 72.3\% & /Z/  & /CH/ & 7.6\%  & 10.0\% \\ \hline
\multirow{3}{*}{Spanish}    
& /UH/ & 72.0\% & 85.0\% & /UW/ & /UW/ & 9.0\%  & 9.3\% \\
& /TH/ & 74.1\% & 39.7\% & /T/  & /T/  & 16.7\% & 44.0\% \\
& /ZH/ & 75.0\% & 20.0\% & /SH/ & /SH/ & 20.0\% & 75.0\% \\ \hline
\multirow{3}{*}{Vietnamese} 
& /ZH/ & 54.2\% & 37.5\% & /S/  & /S/  & 25.0\% & 33.3\%  \\
& /JH/ & 55.6\% & 32.3\% & /CH/ & /CH/ & 10.5\% & 17.7\%  \\
& /AA/ & 58.3\% & 70.5\% & /ER/ & /AO/ & 11.9\% & 11.4\%  \\ \hline
\end{tabular}}
\caption{Summary table of the most often misrecognised target phonemes by the ASR and their most common substitutes including the rates of recognition for each.}
\label{tab:ha_comparison}
\end{table}

The three most frequently misrecognised phonemes for each L1, based on the ASR output, are given in Column 2. Columns 3 and 4 give the recognition rate of each phoneme according to the ASR output (ASR) and the human annotator judgements (HA) respectively. The recognition rate is given as the percentage of occurrences of each phoneme in the prompt texts which were matched in the ASR output or were marked as correct by the human annotator. It can be seen here how the behaviour of the ASR compares to that of the annotators. There are some instances of very similar recognition rates, for example Arabic /TH/ or Korean /TH/. However, more importantly, there are cases where the ASR is either more robust or more sensitive to pronunciation variation compared with human judgements. For instance, Spanish /TH/ is recognised by the ASR 74.1\% of the time whilst the human annotator labels it as correctly produced in only 39.7\% of occurrences. This would suggest that the ASR is relatively robust to the production of /TH/ by Spanish L1 speakers and, whilst its realisation may be considered more /S/ like, it does not have as significant an impact on the recognition accuracy as we might expect. Conversely, consider the case of Hindi /OY/ which the ASR correctly recognises only 57.5\% of the time whilst the human annotator marked 95\% of occurrences as correct. Evidently, the realisation of the /OY/ phoneme by Hindi L1 speakers was not considered different enough from the ``canonical" production as to be labelled a substitution by the human annotator but yet it caused problems with the ASR and contributed to a higher WER for speakers with this L1. 

In the majority of cases, the most commonly substituted phoneme as determined by the ASR output (given in Column 5) matches that identified by the human annotator (given in Column 6). This supports the idea that the behaviour of the ASR is in line with the judgements of a human listener and lends credence to the possibility of leveraging erroneous ASR output to investigate patterns of variation in specific spoken varieties. However, the substitution rates of these Most Common Substitutes (MCS) can vary across L1s and across target phonemes. The substitution rate is given as the percentage of target phoneme occurrences in the prompt text which were recognised as the specific substitute by the ASR or were noted as a substitution with the specific substitute by the human annotator. As would be expected, the substitution rates between the ASR and human annotators have a high degree of variability when the target phoneme recognition rate varies between the two. However, there are also cases where the target phoneme recognition rate is relatively similar from both the ASR and human annotator, but where the substitution rate of the most common substitute differs to a larger extent. For example, the Arabic /TH/ has a recognition rate of 79.2\% by the ASR and 81.0\% by the human annotator and both find /S/ to be the most common substitute. However, the ASR detects this substitution in 7.5\% of /TH/ instances whilst the human annotator notes it in 13.1\% of occurrences. Thus, whilst evaluating an ASR component in this way can reveal a lot of information about its weaknesses and highlight common patterns of phonetic variation across groups of speakers, it does not supply the same phonemic annotation as a human annotator would. 
\section{Discussion}

This sort of sensitivity investigation into the performance of a specific ASR system reveals a lot about how pronunciation variation is handled for different spoken varieties. This can be useful in a number of respects. Firstly, as the use of ASR technology becomes more popular for generating automatic word-level transcriptions of audio data, it is important to understand the limitations of such an approach. Different ASR systems will be trained on different data and will perform differently for different varieties. By testing a specific ASR component on a specific spoken variety or varieties, a better understanding of the expected recognition accuracy can be obtained. ASR systems can be chosen based on their appropriateness for the task and their robustness to style of speech being transcribed. Furthermore, human annotators can be informed of the expected weaknesses of the ASR and the types of errors which are likely to appear within the transcripts thus aiding in the manual correction of automatically generated transcripts. 

From a phonetic variationist approach, the use of a highly sensitive ASR component can reveal systematic pronunciation variation through the patterns of errors which occur in the automatic transcripts. It has been demonstrated that the erroneous output is not random but, in fact, captures commonalities between speakers with similar spoken varieties. Moreover, they are typically aligned with the judgements of human listeners, featuring the same phoneme-level substitutions as identified by the annotator. As such, the investigation presented in this work can aid in highlighting and discovering variation at the phonetic level across speakers from larger and more general accent groups to more specific fine-grained ones depending upon the spoken varieties of the test speakers. Furthermore, the comparison between phonetic variants captured by the ASR output and those detected by human annotators can inform researchers about the capabilities of a particular ASR component in accurately detecting such variants. This is significant information to obtain before incorporating ASR as a tool in variation analysis and annotation. 

This error-based analysis can also be useful for the training and fine-tuning of an ASR system to improve its performance with specific spoken varieties. It enables us to pinpoint specific phonetic realisations which cause issues during recognition. With this information, a targeted approach can be applied to the collection of additional training material, perhaps using prompts specifically designed to elicit those productions which the system is currently unable to handle correctly. Since data collection and the required manual annotation is often time, money and resource consuming, the ability to carry out a smaller scale and more focused collection whilst still improving recognition accuracy is a huge benefit. As previously established, similar error patterns are produced for speakers with similar varieties. Thus, improving the ASR performance for a small subset of test speakers will likely result in improvements for a much greater number of speakers who exhibit similar pronunciation features. Knowledge of the sensitivity or robustness of an ASR system to specific variant pronunciations is vital in developing systems capable of accurately recognising underrepresented and minority varieties.
\section{Conclusion} 

This work has described a method of analysing the output of an ASR system in order to discover how it handles specific phonetic realisations in different L2 Englishes. By examining the output for phoneme substitutions, insertions, and deletions, an awareness is gained of the system's strengths and weaknesses in how it handles pronunciation variation. If a speaker's specific realisation of a phoneme is underrepresented in the ASR training data then it typically leads to the misrecognition of the phoneme which results in higher word error rates. It was demonstrated that such realisations are typically misrecognised in a systematic and consistent way, with the same phoneme substitution recurring throughout the ASR output. Furthermore, the behaviour of the ASR is also systematic across speakers with similar spoken varieties. This was shown through the vector mapping, clustering, and visualisation of the speakers based on the phoneme confusions observed between the read prompt text and the ASR output. Speakers with the same L1 tended to cluster together in the multi-dimensional space because the ASR tended to make the same errors when recognising their spoken utterances. This could be valuable for annotators correcting transcripts initially produced using ASR technology by directing their attention to the errors the system tends to make and what to look out for. This would also suggest that, if these specific productions which the ASR is unable to correctly recognise were to be targeted during fine-tuning, the recognition accuracy of the system would improve both for the sample of speakers tested and for many more speakers with similar varieties. 
Finally, the errors made by the ASR when handling L2 varieties of English were shown to correlate with the judgements of a human annotator. This indicates the feasibility of leveraging ASR to discover and analyse pattern of phonetic variation across groups of speakers. However, given the variability in agreement between the ASR and human annotators with regard to the rate of substitutions, this ASR system would not be suitable for obtaining phoneme level transcripts and should still be considered a tool and aid to manual annotation. 
\renewcommand*{\bibfont}{\footnotesize}
\vspace{\baselineskip}
\bibliographystyle{pwpl_bib}
\bibpunct[:]{(}{)}{,}{a}{}{,}
\bibliography{bib}

\small

\vspace{\baselineskip}
\vspace{\baselineskip}
\noindent ADAPT Research Centre\\
School of Computer Science\\
University College Dublin\\
Dublin, Ireland D04 V1W8\\
\textit{emma.l.oneill@ucdconnect.ie}\\
\textit{julie.berndsen@ucd.ie}
\end{document}